\documentclass[a4paper, 10pt, conference]{ieeeconf}      

\IEEEoverridecommandlockouts                              

\usepackage{graphics} 
\usepackage{epsfig} 
\usepackage{amsmath} 
\usepackage{amssymb}  
\usepackage{bm} 
\usepackage[flushmargin,hang]{footmisc} 

\newcommand{\fig}[1]{Fig.~\ref{#1}}

\newcommand{\eq}[1]{(\ref{#1})}

\def\epsgaiji#1{\leavevmode\kern-0.025zw\raise-.37zh\hbox{%
  \epsfile{file=#1,width=1.05zw}}\kern-0.025zw}
\newcommand{\MARU}[1]{{\ooalign{\hfil#1\/\hfil\crcr\raise.167ex\hbox{\mathhexbox20D}}}}

\usepackage{array}

\usepackage{siunitx} 
\usepackage{gensymb} 

\usepackage{pgfplots}
\pgfplotsset{compat=newest}
\usetikzlibrary{plotmarks}
\usetikzlibrary{arrows.meta}
\usepgfplotslibrary{patchplots}
\usepackage{grffile}
\pgfplotsset{plot coordinates/math parser=false}
\newlength\fwidth
\newlength\fheight

\usepackage{cite}

\title{\LARGE \bf
Lower Gravity Demonstratable Testbed for Space Robot Experiments}

\author{Kentaro Uno, Kazuki Takada, Keita Nagaoka, Takuya Kato, Arthur Candalot, and Kazuya Yoshida
    \thanks{This work is supported by JSPS KAKENHI Grant Number JP23K13281. The authors also would like to thank Hiroki Nishikori, Louis Mamelle, Taku Okawara and Keigo Haji for their valuable contributions.}%
    \thanks{
    K. Uno, K. Takada, K. Nagaoka, T. Kato, A. Candalot, and K. Yoshida are with the Space Robotics Lab. (SRL) in Department of Aerospace Engineering, Graduate School of Engineering, Tohoku University, Sendai 980--8579, Japan. (E-mail: \tt{unoken@tohoku.ac.jp}) 
    }%
    \thanks{
    \textit{*The corresponding author is Kentaro Uno.}
    }%
}%

\begin{document}

\maketitle
\thispagestyle{empty}
\pagestyle{empty}


\begin{abstract}
In developing mobile robots for exploration on the planetary surface, it is crucial to evaluate their performance, demonstrating the harsh environment in which the robot will actually be deployed. Repeated experiments in a controlled testing environment reproducing various terrain and gravitational conditions are essential. This paper presents the development of a minimal and space-saving indoor testbed, which can simulate steep slopes, uneven terrain, and lower gravity, employing a three-dimensional target tracking mechanism (active xy and passive z) with a counterweight. 
\end{abstract}



\section{INTRODUCTION}
Space robots work under different gravity (Moon: 1/6\;G, Mars: 3/8\;G, and small bodies or in-orbit: micro\;G). Thus, it is essential to have an experimental setup that can demonstrate that the robot moves under such a lower gravity. Gravity offloading techniques for such a space robot experiment are generally divided into three types: 1) Air-floating: This method removes the effect of gravity on motion in a two-dimensional plane by levitating the object on the surface using compressed air and reducing friction to as close to zero as possible~\cite{yuguchi2016verification}. By changing the tilt angle of the surface plate, arbitrary gravity can be simulated. 2) Applying a controlled external force: This technique utilizes a robot arm~\cite{chacin2008microgravity} or a tether~\cite{valle2017reduced} attached to the target object to externally apply the controlled forces and moments other than gravity with torque and force sensor feedback, resulting in that any different gravitational forces and external moments can be simulated for the target. In this technique, we can mimic any gravity with high accuracy, but the whole hardware system tends to be large-sized. 3) Counterweighting: In this technique, the target object is hoisted by a tether to which a counterweight is attached at the other end. A vertical upward force is applied to the centroid of the object~\cite{brown1994novel}. Changing the counterweight ratio to the object's weight can simulate an arbitrary gravity environment. The third method is less accurate than the second solution because the force is not actively controlled. However, the mechanical setup and control are more straightforward.

\begin{figure}[t]
  \centering
  \includegraphics[width=\linewidth]{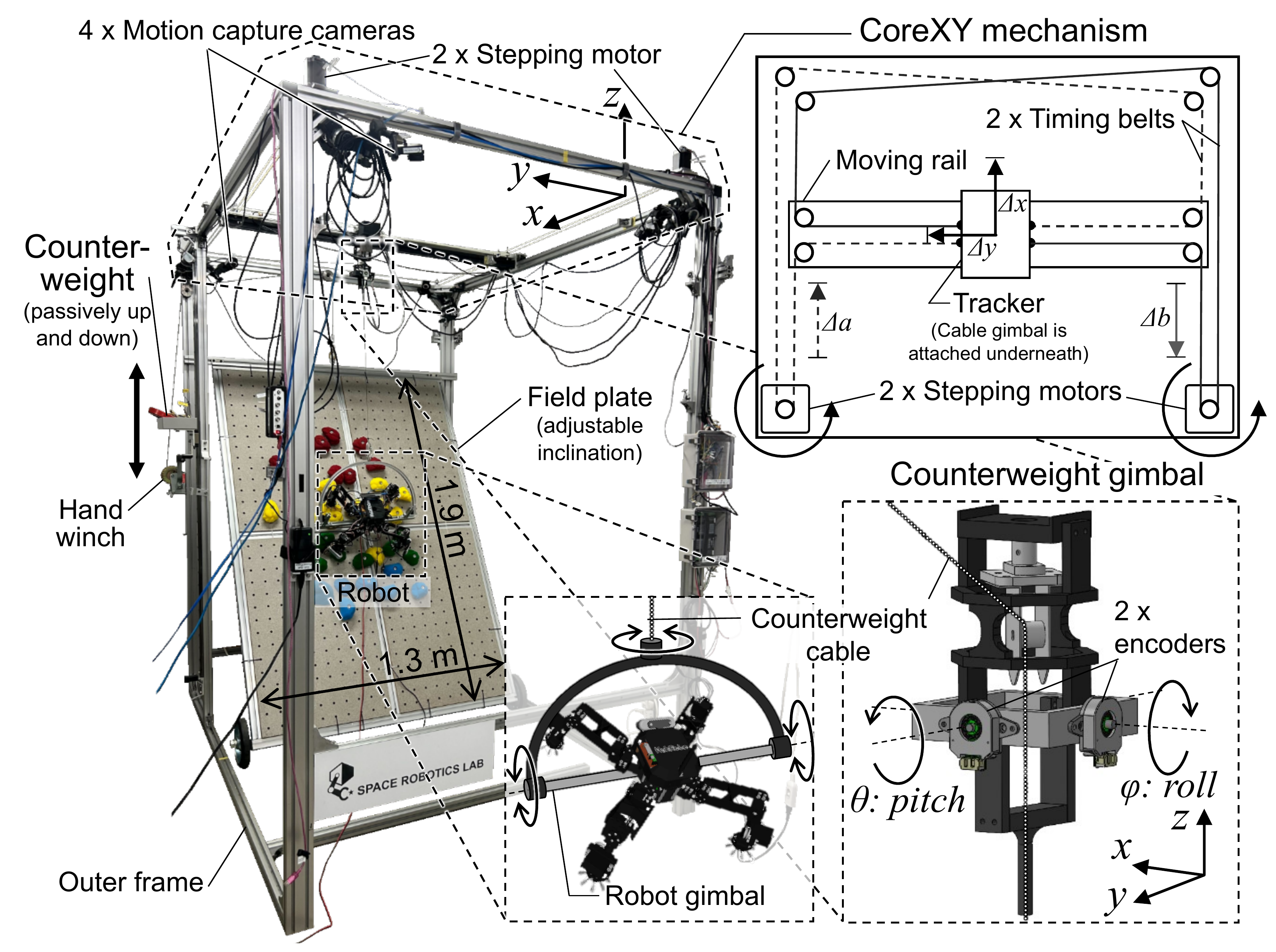}\vspace{-2mm}
  \caption{Developed terrain-, inclination-, and gravity-adjustable testbed. The hoisted robot's mass is partially (or fully) compensated by a counterweight. The cable of the counterweight is maintained to be vertical while the robot moves on the field using the CoreXY mechanism.}
  \vspace{-2mm}
  \label{testfield}
\end{figure}

This paper focuses on developing the indoor testbed, which can change the inclination angle of the field and reproduce various uneven terrains for space robotic experiments. The entire system is realized in a cost- and space-saving manner by means of the counterweighting solution for the gravity compensation and integration of the CoreXY mechanism~\cite{corexy} on the ceilings of the outer frame. This mechanism is used to actively track the moving robot's horizontal position, which helps sustain the cable of the counterweight (i.e., the direction of the gravity offload) always vertical while the robot moves.


\section{SYSTEM DESIGN AND INTEGRATION}
The developed test field consists of an outer frame, a field plate that can change its inclination along with the outer frame (see \fig{testfield}), and a three-dimensional (3D) target-tracking mechanistic system on the outer cage. The field plate frame can accommodate four wooden top panels, which have holes at 60\;mm intervals to attach any object (maximum payload capacity: 200\;kg) such as steps, handholds, rocks, or sand tray, creating any artificial or natural rocky/sandy terrain assuming the in-orbit space station interior or the planetary surface. The field plate's inclination angle can easily vary by the experimenter's turning the attached hand-cranked winch that lifts the backside edge of the plate. In this testbench, the robot is hung by a cable pulled by a counterweight; the cable is fixed to the robot via a two degree-of-freedom (pitch and yaw) free rotating gimbal, which helps not to give an undesired moment to the robot.

The testbench has a 3D target tracking system to keep applying a vertical gravity offloading force at the moving robot's center of mass. In this system, $z$-position is passively adjusted by the vertically moving counterweight on the friction-less rail, and $xy$-position is actively tracked using CoreXY mechanism~\cite{corexy}, which is typically used in 3D printer nozzle's mobility. 
CoreXY mechanism enables the precise control of $xy$-position of the tracker by conveying the belts, which are driven by the two stepping motors. 
The tracker's necessary displacement to set the cable's tilt angle back to zero is derived as follows.
Let {$l$} be the length of the cable from the robot fixture to the tracker, and {$\phi$} and {$\theta$} be the roll and pitch angles of the cable from the vertical state relative to the inertial frame, respectively. The cable's roll and pitch tilt angle are measured by two encoders attached to the upper gimbal through which the counterweight cable is. The amount of movement in the {$x$}- and {$y$}-axes {$\Delta x$} and {$\Delta y$} can be obtained as Eq.~\eq{xy}.
\begin{align}
\Delta x = - l \sin \theta, \ \Delta y = l \sin \phi
\label{xy}
\end{align}
In the CoreXY schematic of \fig{testfield}, if the two belt feeds driven by the left and right stepping motors are {$\Delta a$} and {$\Delta b$}, respectively, and these can be obtained from {$\Delta x$} and {$\Delta y$} as computed in Eq.~\eq{ab}.
\begin{align}
\Delta a = \Delta x - \Delta y, \ \Delta b = - \Delta x - \Delta y 
\label{ab}
\end{align}
Two timing belt feeds are determined using Eq.~\eq{xy} and Eq.~\eq{ab}. The control input of the stepping motors is determined by PID control. The data is logged by ROS software. 





\section{EVALUATION}
\begin{figure}[t]
  \centering
    \includegraphics[width=.9\linewidth,clip]{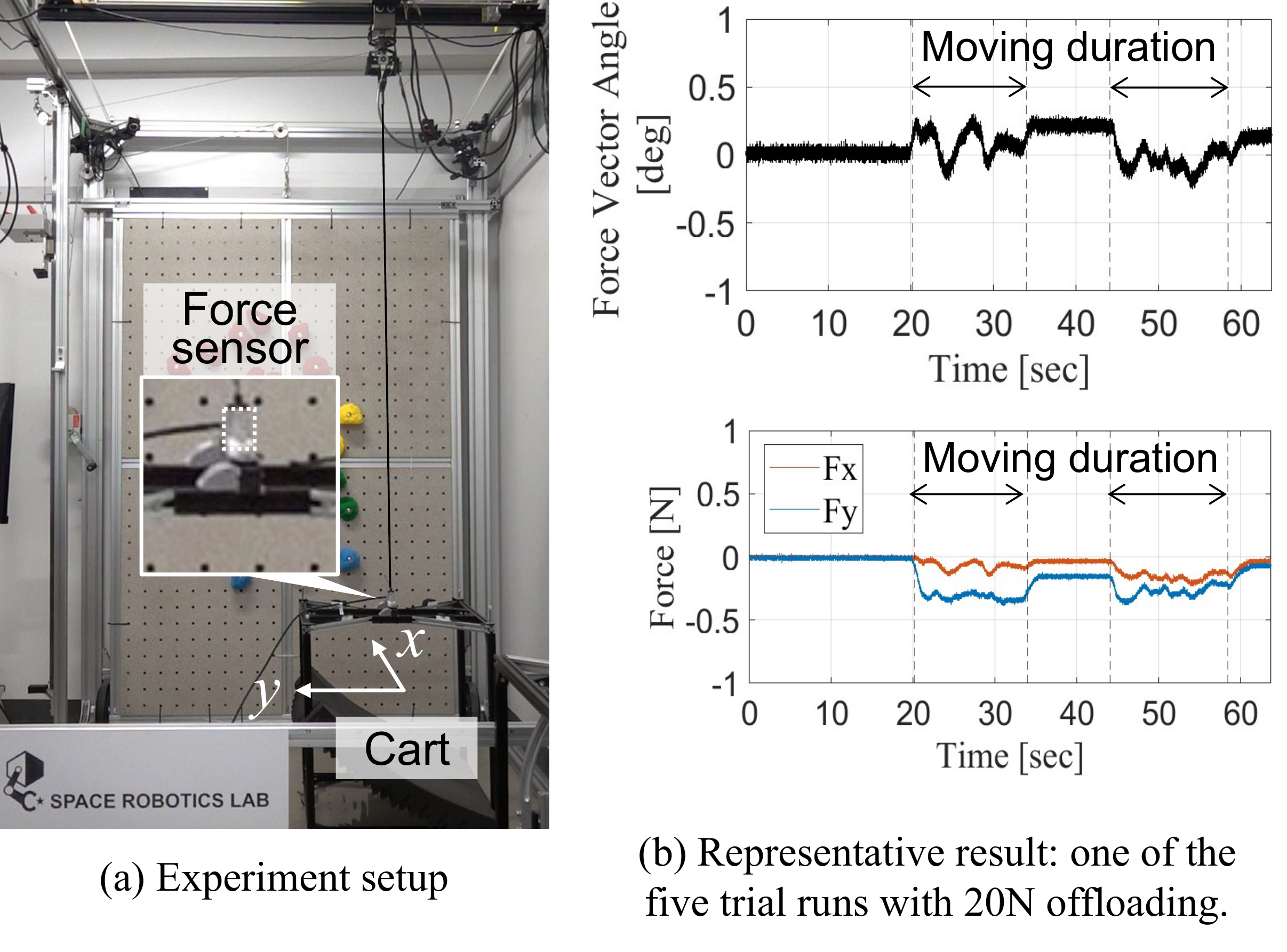}
  \caption{System performance experiment: (a) Test setup and (b) Angle from the vertical (upside plot) and horizontal components of the gravity offloading force (upside plot) acting on the target.}
  \vspace{-2mm}
  \label{testResult}
\end{figure}
In our developed test bench, the target tracking system keeps the counterweighting force precisely vertical along with the robot movement, which is essential to prevent the robot from getting unnecessary horizontal disturbance by the controlled force for gravity cancellation. 
An evaluation test was conducted to assess the performance of the system. For this test, a cart was attached instead of a robot, and a three-axis force sensor was inserted between the cable and the cart's top surface to measure the counter force's vector (see \fig{testResult}(a)). The cart was linearly almost 1\;m pushed towards the direction of positive $x$ and $y$, and the force vector and the cart velocity were recorded by the force sensor and the motion capturing system, respectively. The compensation force value was set to 20\;N or 40\;N, and the test was conducted five times, respectively. A representative result is shown in \fig{testResult}. Through the total ten attempts, the average velocity of the cart was 34\;mm/s (min: 25\;mm/s, max: 43\;mm/s). Throughout all trials, while the target moves, the angle from the vertical of the force acting on the target was kept between {$\pm 1.5$\;deg} (\fig{testResult}(b)), which results in horizontal components of the force being less than only 0.5\;N (\fig{testResult}(c)), which is 1.3\% of the 40\;N of the offloading force.
\begin{figure}[t]
  \centering
  \includegraphics[width=\linewidth]{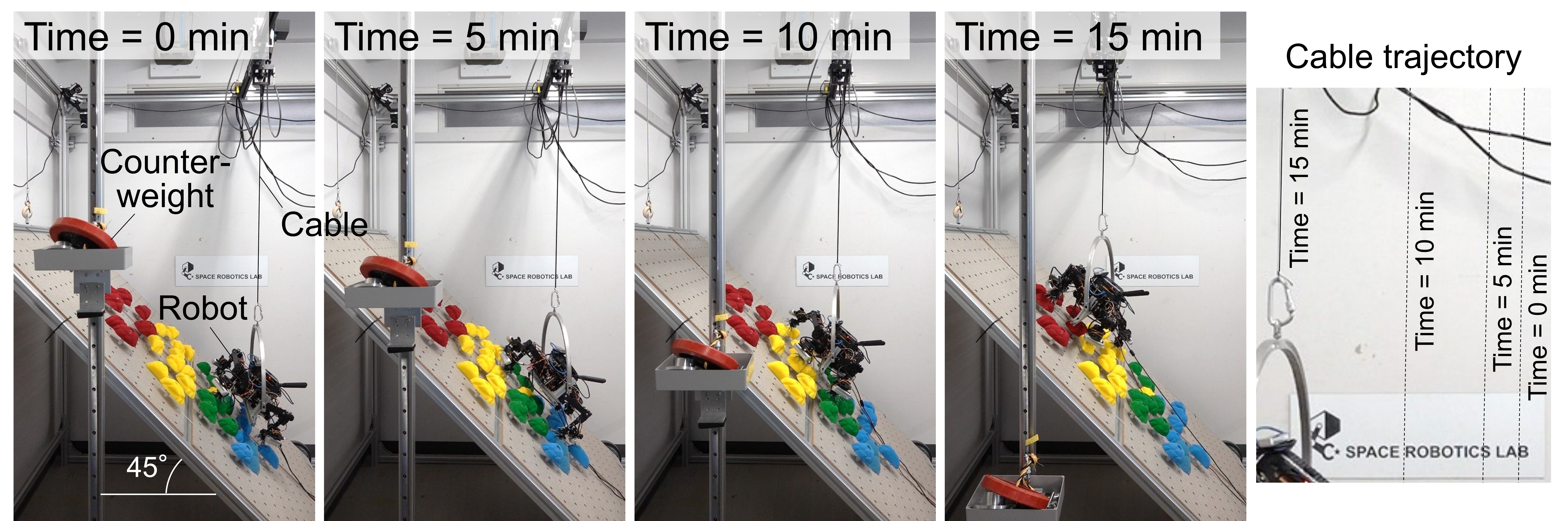}
  \caption{Space exploring climbing robot experiment with demonstrating the gravity on the Moon.}
  \vspace{-2mm}
  \label{climbing}
\end{figure}
Furthermore, a space exploration robot demonstration using the testbed is showcased in \fig{climbing}. In this demonstration, a four-limbed cliff-climbing robot, HubRobo \cite{uno2021hubrobo} (mass: 3.0\;kg), was deployed on the 45$^\circ$ inclined irregular slope. The counterweight mass was determined to be 3.5\;kg (robot gimbal mass (1.0\;kg) + 5/6 of HubRobo mass), simulating the Lunar gravity. In practice, a certain percentage of the calculated value was added to the counterweight to offset the friction. By the accurate target tracking, we confirmed the force for gravity offloading was kept vertical (see the cable trajectory in \fig{climbing}) during the robot's locomotion.




\section{CONCLUSION}
This paper detailed the development of an experimental environment for space robots in which the terrain unevenness, slope inclination, and simulated gravity can be adjusted. In this testbed, a 3D target tracking system (active $xy$ and passive $z$ adjustment) was constructed to keep applying a counter gravity vertical force for the moving target, and the performance of this system was confirmed to be satisfactory through the experiments. 
Future scope includes achieving more accuracy of the simulated gravity by adding a $z$-axis linear actuator and a force sensor to the system, which completes an active 3D control with force feedback.



\bibliography{./IEEEabrv,reference}

\begin{thebibliography}{1}
\providecommand{\url}[1]{#1}
\csname url@rmstyle\endcsname
\providecommand{\newblock}{\relax}
\providecommand{\bibinfo}[2]{#2}
\providecommand\BIBentrySTDinterwordspacing{\spaceskip=0pt\relax}
\providecommand\BIBentryALTinterwordstretchfactor{4}
\providecommand\BIBentryALTinterwordspacing{\spaceskip=\fontdimen2\font plus
\BIBentryALTinterwordstretchfactor\fontdimen3\font minus
  \fontdimen4\font\relax}
\providecommand\BIBforeignlanguage[2]{{%
\expandafter\ifx\csname l@#1\endcsname\relax
\typeout{** WARNING: IEEEtran.bst: No hyphenation pattern has been}%
\typeout{** loaded for the language `#1'. Using the pattern for}%
\typeout{** the default language instead.}%
\else
\language=\csname l@#1\endcsname
\fi
#2}}

\bibitem{yuguchi2016verification}
Y.~Yuguchi \emph{et~al.}, ``Verification of gait control based on reaction
  null-space for ground-gripping robot in microgravity,'' in \emph{Proc. IEEE
  ICRA}, 2016, pp. 2822--2827.

\bibitem{chacin2008microgravity}
M.~Chacin \emph{et~al.}, ``A microgravity emulation testbed for asteroid
  exploration robots,'' in \emph{Proc. i-SAIRAS}, 2008.

\bibitem{valle2017reduced}
P.~Valle, ``Reduced gravity testing of robots (and humans) using the active
  response gravity offload system,'' in \emph{Proc. IEEE/RSJ IROS}, 2017.

\bibitem{brown1994novel}
H.~B. Brown \emph{et~al.}, ``A novel gravity compensation system for space
  robots,'' in \emph{ASCE Specialty Conference on Robotics for Challenging
  Environments}, 1994, pp. 250--258.

\bibitem{corexy}
I.~E. Moyer, ``Core xy,'' 2012, \url{https://corexy.com/reference.html
  (Accessed: Aug. 8th, 2023)}.

\bibitem{uno2021hubrobo}
K.~Uno \emph{et~al.}, ``Hub{R}obo: A lightweight multi-limbed climbing robot
  for exploration in challenging terrain,'' in \emph{Proc. IEEE-RAS Humanoids},
  2021, pp. 209--215.

\end{thebibliography}

\end{document}